\documentclass[conference]{IEEEtran}
\IEEEoverridecommandlockouts
\usepackage{cite}
\usepackage{amsmath,amssymb,amsfonts}
\usepackage{algorithmic}
\usepackage{graphicx}
\usepackage{textcomp}
\usepackage[table]{xcolor}
\def\BibTeX{{\rm B\kern-.05em{\sc i\kern-.025em b}\kern-.08em
    T\kern-.1667em\lower.7ex\hbox{E}\kern-.125emX}}
\usepackage{orcidlink}    
\usepackage{booktabs,multirow,bigstrut}
\usepackage[nameinlink,noabbrev,capitalise]{cleveref}

\begin{document}

\title{Architectural Exploration of Hybrid Neural Decoders for Neuromorphic Implantable BMI
\thanks{
This work was partially supported by the National Research Foundation, Prime Minister’s Office, Singapore under its Campus for Research Excellence and Technological Enterprise (CREATE) - IN-CYPHER program, and the Research Grants Council of the HK SAR, China (Project No. CityU 11200922).
}
}

\author{ 
Vivek~Mohan\textsuperscript{*\textdaggerdbl}~\orcidlink{0000-0002-0248-6417},
Biyan Zhou\textsuperscript{\textdagger}~\orcidlink{0009-0003-7725-8974},
Zhou Wang\textsuperscript{*\textdaggerdbl}~\orcidlink{0000-0002-0203-6854},
Anil Bharath\textsuperscript{*}~\orcidlink{0000-0001-8808-2714},
Emmanuel Drakakis\textsuperscript{*}~\orcidlink{0000-0001-6649-0573},
Arindam~Basu\textsuperscript{\textdagger}~\orcidlink{0000-0003-1035-8770}\\ \textsuperscript{*}\textit{Imperial College London}~\textsuperscript{\textdaggerdbl}\textit{Imperial Global Singapore}~\textsuperscript{\textdagger}\textit{City University of Hong Kong}\\
Email: v.mohan@imperial.ac.uk,
arinbasu@cityu.edu.hk
}

\maketitle

\begin{abstract}
This work presents an efficient decoding pipeline for neuromorphic implantable brain-machine interfaces (Neu-iBMI), leveraging sparse neural event data from an event-based neural sensing scheme. We introduce a tunable event filter (EvFilter), which also functions as a spike detector (EvFilter-SPD), significantly reducing the number of events processed for decoding by $192\times$ and $554\times$, respectively. The proposed pipeline achieves high decoding performance, up to $\mathrm{R^2}=0.73$, with ANN- and SNN-based decoders, eliminating the need for signal recovery, spike detection, or sorting, commonly performed in conventional iBMI systems. The SNN-Decoder reduces computations and memory required by $5-23\times$ compared to NN-, and LSTM-Decoders, while the ST-NN-Decoder delivers similar performance to an LSTM-Decoder requiring $2.5\times$ fewer resources. This streamlined approach significantly reduces computational and memory demands, making it ideal for low-power, on-implant, or wearable iBMIs.
\end{abstract}


\section{Introduction}
Advances in implantable brain machine interface (iBMI) based neurotechnology have enabled the realization of therapies for epilepsy via deep brain stimulation and mental disorders \cite{MentalBMI}, and potentially partial restoration of lost sensory \cite{VisionBMI}\cite{speechBMI2023} and motor capabilities \cite{Neuroprosthetic}.
A typical implementation of an iBMI system involves recording neural activity through a microelectrode array followed by stages of amplification, filtering, and spike-detection to capture the action potentials which then may be decoded to control effectors such as a prosthetic arm, computer cursor, mobility device, etc.
While most existing iBMI systems comprise wired low-electrode count implants, next-generation iBMIs (Nx-iBMI) are expected to be wireless and capable of simultaneously recording from thousands of neurons, enhancing iBMI performance and functionality \cite{naturePowerSaving}.
Event-based or neuromorphic iBMIs (Neu-iBMI) have been proposed in \cite{NeuromorphicRecSystem, NSS_PNS, ISCAS2023, NCE2025, He2024, 4096ch_eventbasedMEA} to address the power, bandwidth, and implant size limitations that arise with increasing the number of recording electrodes.

Several linear and non-linear decoding algorithms have been proposed for a variety of iBMI tasks such as control of a prosthetic or robotic arm, hand kinematic or speech decoding, etc. Traditional decoding algorithms such as linear regression \cite{Collinger2013_prostheticArm}, Kalman filter (KF) \cite{ibmi_kf_1, ibmi_kf_2, spikingKF_ibmi} and linear discriminant analysis \cite{RosaSoPlos}, are examples of linear decoders. With the rapid growth of artificial neural networks (ANN), non-linear decoders using recurrent neural networks (RNN) and long-short term memory (LSTM), are actively being studied due to their high speed and high performance \cite{RNN_ibmi, Brain2Text, Ahmadi_LSTM, Collinger2013_prostheticArm}. Works in \cite{shoeb_RL, shallowNN_ibmi, SNN_ibmi_decoder} introduce some neural network based decoders that either incorporate neuromorphic principles or SNNs in the algorithms. All of these decoders use binned firing rates or other features extracted from the filtered neural signal, digitized at the Nyquist sampling rate. While decoders of varying degrees of complexity can be implemented on a general-purpose computer, a wireless Nx-iBMI system necessitates the exploration of lightweight shallow neural network decoders running directly on the implant or a connected wearable.
 While studies in \cite{NeuromorphicRecSystem, NSS_PNS, 4096ch_eventbasedMEA} demonstrate the feasibility of integrating Neu-iBMIs with SNN processors such as DYNAP-SE \cite{DYNAP} and \cite{ISCAS2024} introduces hybrid event-frame spike detection methods for Neu-iBMIs, they do not explore decoding schemes that utilize the event-based neural data from these Neu-iBMIs for realistic iBMI tasks such as motor/intention decoding.

This work fills the void in the area of event-based neural decoding for Neu-iBMIs and presents an architectural exploration of hybrid neural decoders that operate directly on the event stream generated by the Neuromorphic Compression based Neural Sensing (NCNS) scheme introduced in \cite{ISCAS2023, NCE2025} as shown in Fig. \ref{fig:BMI_Decoding_pipeline}(a). We make the following contributions:
\begin{itemize}
\item An event filtering scheme for NCNS that can be tuned to either suppress events generated by background or operate as a spike detector co-designed in the decoder to detect potential spikes and can reduce the number of events to be processed by $\mathrm{192-544\times}$. Its output is then fed to the feature extraction block of the decoder.
\item An architectural exploration of hybrid decoders--- two shallow MLP-based neural networks and an event-based SNN decoder (requiring $5-23\times$ fewer resources than NN-based decoders) for decoding hand-reaching tasks performed by a non-human primate (NHP) directly from NCNS generated neural events, resulting in $\mathrm{R^2}$ up to $0.73$.
\end{itemize}

\section{Methodology}\vspace{-0.1cm}
\subsection{Event Filter}
\begin{figure}[t!]
    \centering
    \includegraphics[width=0.48\textwidth]{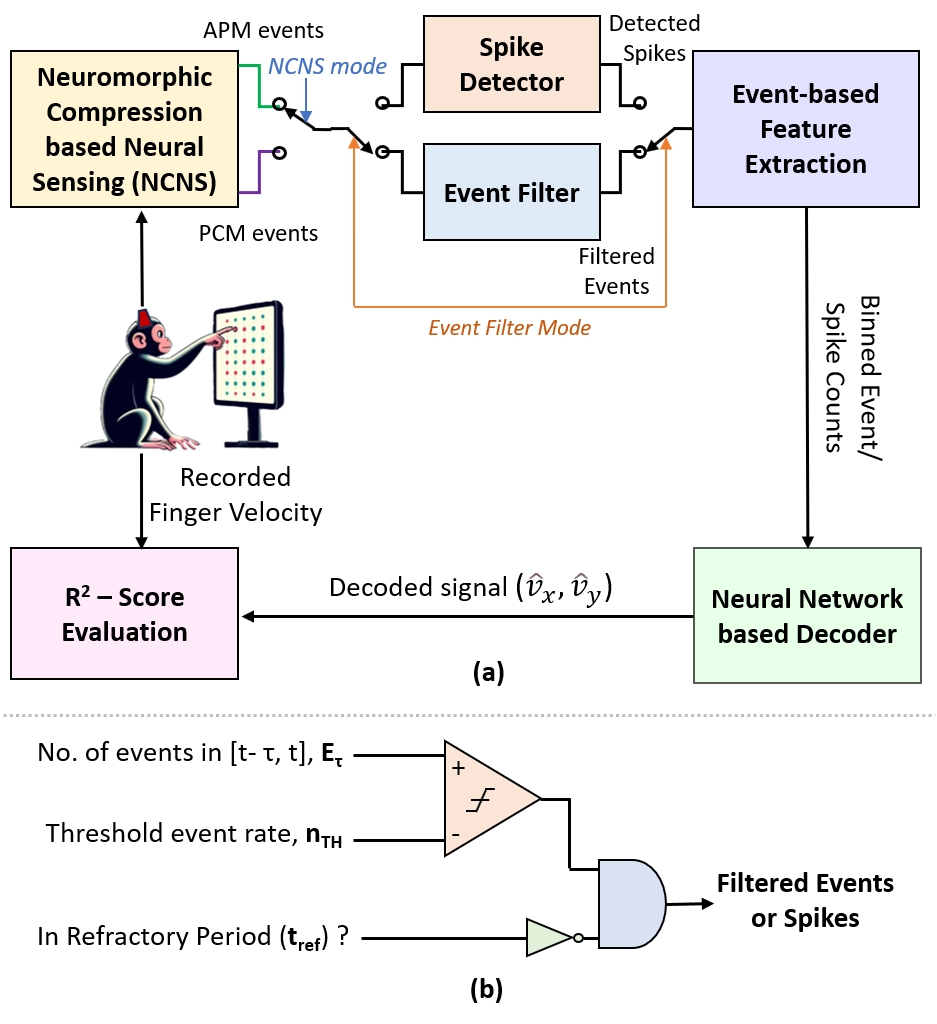}
    \caption{(a) Neural decoding pipeline for NCNS-based (\cite{ISCAS2023, NCE2025}) Neu-iBMI. Events from NCNS are passed through a spike detector or event filter, which allows events corresponding to the spike. Event-based feature extraction involves binning events for some ANN-based decoder models, while filtered event data or spikes may be directly streamed through SNN-Decoder. $\mathrm{R^2}$ score is the performance metric used. (b) Logical block diagram of an event filter for NCNS background activity suppression. A spike detector can be realized as a special case of the event filter by choosing an appropriate period of refraction.}
    \vspace{-0.4cm}
    \label{fig:BMI_Decoding_pipeline}
\end{figure}
Taking into account the nature of event generation in NCNS, a BAF/STCF (for DVS) \cite{dvs_tobi} inspired \emph{EvFilter} (shown in Fig. \ref{fig:BMI_Decoding_pipeline}(b)) allows an event $e_i$ to pass through it, if there have been at least $\mathrm{n_{th}}$ past events ($E_{\tau}$) in a temporal neighborhood $\tau$ of the incoming event for the recording channel. The EvFilter may include a refractory period $\mathrm{t_{ref}}$ for which no events are allowed to pass through once a spike event is detected, to prevent spurious events (from scenarios like flickering noise) from passing through. EvFilter with $\mathrm{t_{ref}} = 0$ allows all events potentially corresponding to the spike to pass through.
A spike detector, \emph{EvFilter-SPD}, can be realized as a special case of the EvFilter where the refractory period $\mathrm{t_{ref}}$ is set large enough (typically $1$ ms) to allow only one detection per spike.
\vspace{-0.1cm}

\subsection{Neural Decoders for Neuromorphic iBMI}
Binned spikes (rate of threshold-crossing based spike detection) are a popular feature in BMI decoding \cite{ChenYi_elm, ShoebiBMI, shoeb_RL, Ahmadi_LSTM}. In this work, for the NCNS iBMI decoding, binning is implemented after EvFilter or EvFilter-SPD, effectively representing event counts per bin period. The accumulated events in fixed bins are in essence a temporal frame capturing the dynamics of the neural signal.
\subsubsection{NN-Decoder}
\begin{figure} [t]
\centering    
\includegraphics[width=0.48\textwidth]{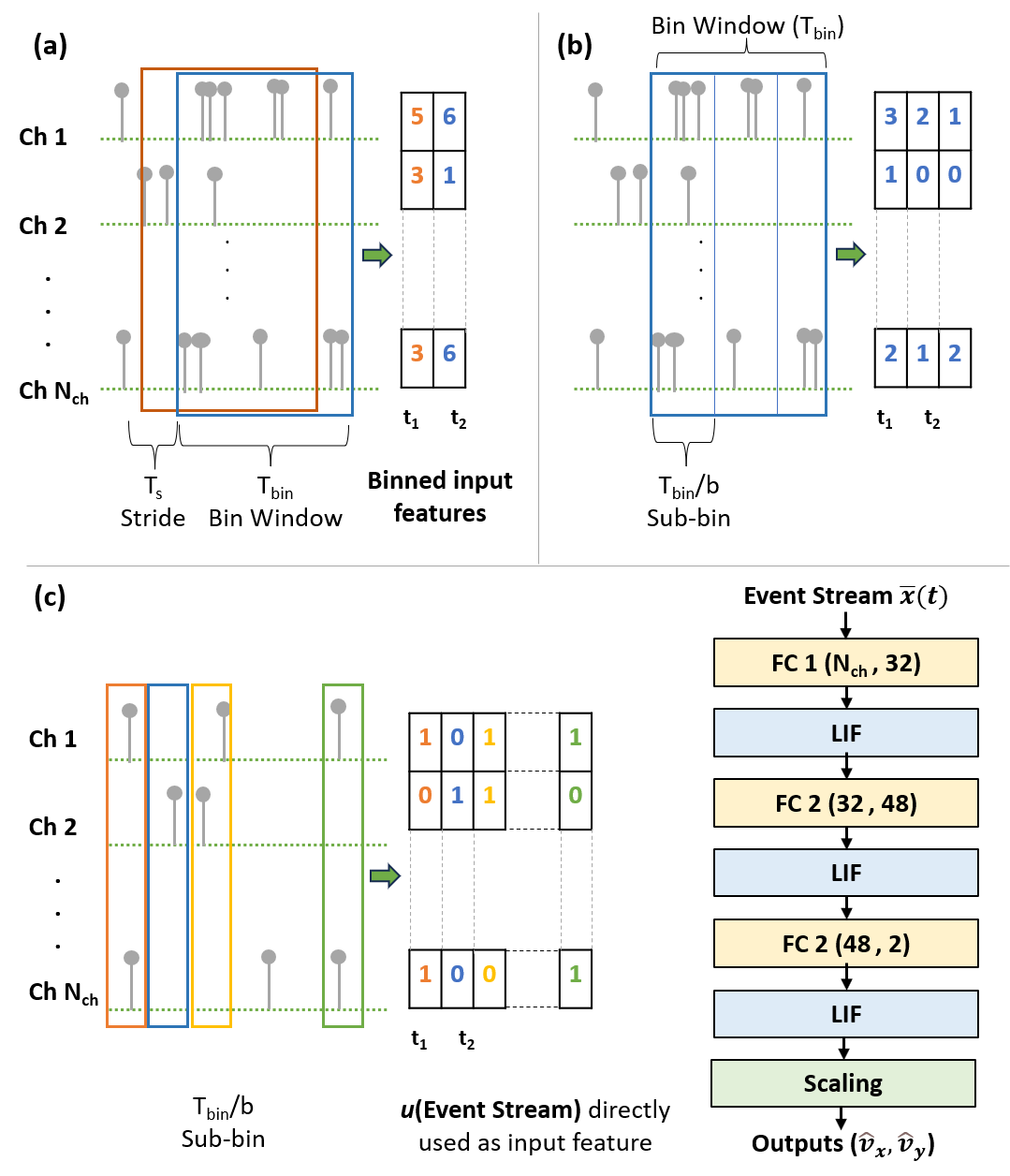}
 \caption{Input feature extraction for (a) NN-Decoder and (b) ST-NN-Decoder (c) SNN-Decoder (right: network architecture).}
\label{fig:NN_Decoder}
\vspace{-0.5cm}
\end{figure}

    Filtered events after EvFilter or detected spikes after EvFilter-SPD are accumulated into bins of duration $\mathrm{T_{bin}}$ at sample time $t$, with a stride of $\mathrm{T_{s}}$ which corresponds to the sampling duration of the output hand kinematic as shown in Fig. \ref{fig:NN_Decoder}(a). The accumulated bins $\mathrm{\bar{x}_{NN}}(t)$ across $\mathrm{N_{ch}}$ recording channels form the input feature vector to the NN-Decoder,  represented as:
    \begin{align}\label{eq:binning}
         \mathrm{\bar{x}_{NN}}(t) = [\mathrm{x_1}(t),~\mathrm{x_2}(t),~...,~\mathrm{x_{N_{ch}}}(t)],
    \end{align}
    where, $\mathrm{x}_i(t)$ holds the event or spike counts for the $i^\mathrm{th}$ recording channel.
    The network consists of fully connected layers, arranged as - input layer of $\mathrm{N_{ch}}$ neurons, hidden layer $1$ and $2$ with $32$ and $48$ layers (informed from exploration in \cite{Biyan_iBMI_ANN_SNN}), respectively, and two output neurons corresponding to the x and y velocities for hand kinematic decoding task. Rectified linear unit (ReLU) is used as the activation function for the first two layers and, batch normalization and dropout ($50\%$ probability) are implemented to improve accuracy and generalize the model.

\subsubsection{Segmented Time-bins NN-Decoder (ST-NN-Decoder)}
For ST-NN-Decoder, binning is done at finer temporal intervals by segmenting the bin window into $b$ bins, each of duration $\mathrm{T_{bin}}/b$ units. Filtered events after EvFilter or detected spikes after EvFilter-SPD are therefore accumulated into $b$ sub-bins of duration $\mathrm{T_{bin}}/b$. The channel-wise bin counts, which collectively form the feature vector, are now vectors containing the bin counts from each of the sub-bins. The bin counts for the $i^\mathrm{th}$ recording channel at sample time $t$ is given as:
    \begin{align}
        \mathrm{\bar{x}}_i(t) = [\mathrm{c}^{1}_i(t), ~\mathrm{c}^{2}_i(t),~..., ~\mathrm{c}^{b}_i(t)],
    \end{align}
    where, $\mathrm{c}^{k}_i(t)$ represents the count accumulated in the $k^\mathrm{th}$-bin of the $i^\mathrm{th}$-channel.
   The accumulated counts in the segmented bins across $\mathrm{N_{ch}}$ recording channels form the input feature vector to the ST-NN-Decoder, represented as:
    \begin{align}\label{eq:st_binning}
         \mathrm{\bar{x}_{ST-NN}}(t) = [\mathrm{\bar{x}_1}(t),~ \mathrm{\bar{x}_2}(t), ~..., ~\mathrm{\bar{x}_{N_{ch}}}(t)],
    \end{align}
    The input feature vector made up of $b$-dimensional counts from $b$ of $\mathrm{N_{ch}}$ recordings is subsequently flattened to form $\mathrm{N_{ch}} \times b$ feature vector as shown in Fig. \ref{fig:NN_Decoder}(b).
    The network for ST-NN-Decoder is similar to NN-Decoder except that input layer now consists of $\mathrm{N_{ch}}\times b$ neurons.

\subsubsection{SNN-Decoder}
    This is implemented to process the incoming events or spikes as a continuous stream without additional binning for feature extraction. The input feature vector for SNN-decoder $\mathrm{\bar{x}_{SNN}}(t)$ at sample time $t$ is a binary vector, mathematically represented as:
    \begin{align}\label{eq:snn_binning}
         \mathrm{\bar{x}_{SNN}}(t) = [u(\mathrm{x}_1(t)), ~u(\mathrm{x}_2(t)), ~ ..., ~u(\mathrm{x_{N_{ch}}}(t))],
    \end{align}
    where $u()$ is a unit-step or Heaviside function applied to the $i_\mathrm{th}$ channel, encoding the absence or presence of a spike in each of the $\mathrm{N_{ch}}$ channels at sample time $t$ as shown in Fig. \ref{fig:NN_Decoder}(c)(left). This kind of binary encoding enables the implementation of SNN with selective accumulation (AC) instead of using multiply and accumulate (MAC) operations.
    The inherently recurrent and stateful nature of the neurons containing memory in SNNs make them a good candidate for iBMI decoding. The SNN-Decoder implemented in this work as shown in Fig. \ref{fig:NN_Decoder}(c)(right), consists of three fully-connected (FC) layers with leaky integrate-and-fire (LIF) neurons in each layer. The behavioral model of the LIF neurons is captured in the following set of equations \cite{Biyan_iBMI_ANN_SNN}:
    \begin{align}\label{eq:lif}
        \mathrm{U}[t] &= \beta \mathrm{U}[t-1] + \mathrm{WX}[t] - \mathrm{S_{out}}[t-1]\theta \notag\\
        \beta &= e^{-\Delta t/\tau} \notag\\
        \mathrm{S_{out}} &= \begin{cases}
        1, &\text{if}~\mathrm{U}[t]>\mathrm{U_{thr}}\\
        0, &\text{otherwise}
        \end{cases} \notag \\
        \theta &= \begin{cases}
        0, &\text{if~no~reset}\\
        \beta\mathrm{U}[t-1]+\mathrm{WX}[t], &\text{if~reset-to-zero}
        \end{cases}
    \end{align}
$\mathrm{X}[t]$ is the input and $\mathrm{U}[t]$ in Eq. \ref{eq:lif} is the membrane potential of the LIF neuron with threshold membrane potential $\mathrm{U_{thr}}$ at the $t^\mathrm{th}$ time step. The synaptic weights and decay rate of the FC-layers are $\mathrm{W}$ and $\beta$, respectively, and the reset mechanism is defined by $\theta$, with a time step interval of $\Delta t$. Each layer of the SNN-Decoder has its separate $\mathrm{U_{thr}}$ and $\beta$ (learned during training). The first and second FC-layers are implemented with a reset-to-zero mechanism, whereas, no reset is implemented for the last layer to allow for the accumulation of membrane potential. The outputs $\mathrm{V_x}$ and $\mathrm{V_y}$ corresponding to the subject's movements predicted by the SNN-Decoder are determined by scaling the accumulated membrane potential of neurons in the last layer with a learnable constant parameter.

\section{Results}
\subsection{Dataset}
This work utilizes the primate reaching dataset made available in \cite{NHP_HandReaching_dataset} which provides the 96-channel raw baseband recordings and corresponding spike times (`GT - Spike Times'). Five recordings including the first and last day of trials for the NHP `Indy' were converted to neural events following the NCNS simulation pipeline presented in \cite{ISCAS2023, NCE2025} and then processed using EvFilter/EvFilter-SPD.
For decoder training, the time series data was segmented into reaches by detecting changes in the target position provided in the dataset, followed by splitting them into training ($50\%$), validation ($25\%$), and test ($25\%$) sets according to the number of reaches (similar to \cite{Neurobench, Shoeb_Selma, Biyan_iBMI_ANN_SNN}).

\subsection{Evaluation Metrics}
The coefficient of determination, commonly known as $\mathrm{R^2}$, serves as a crucial evaluation metric in decoding tasks such as predicting hand kinematics from neural signals \cite{ShoebiBMI, Neurobench, NHP_HandReaching_dataset}. If ${\mathrm{Y}} = \{y_1,~ y_2,~ ...,~ y_n\}$ is a vector containing the expected or target values with mean value $\bar{y}$, and $\hat{\mathrm{Y}} = \{\hat{y}_1,~ \hat{y}_2,~ ...,~ \hat{y}_n\}$ contains the predicted values, then the $\mathrm{R^2}$-score for $n$ observations is determined as follows:
\begin{align}
    \mathrm{R^2} = 1 - \frac{\sum_{i=1}^{n} (y_i - \hat{y}_i)^2}{\sum_{i=1}^{n} (y_i - \bar{y})^2}
\end{align}
$\mathrm{R^2}$ values range from 0 to 1, where a higher score indicates superior predictive performance, signifying the model's ability to accurately capture the relationship between recorded neural data and observed hand movements. For NHP hand reaching prediction done in this work, $\mathrm{R^2_x}$ and $\mathrm{R^2_y}$, corresponding to x- and y-velocity prediction are determined separately, and the overall metric is obtained by computing their mean: $\mathrm{R^2} = (\mathrm{R^2_x}+\mathrm{R^2_y})/2$.

\subsection{Decoding Performance}
\subsubsection{Decoder Model Training and Testing}
All the models presented in this work were trained for 50 epochs using the SNNTorch framework \cite{snnTorch_Paper} with learning rate of $0.005$, a dropout rate between $0.3-0.5$, using L2-regularization value between $0.005-0.2$. AdamW was chosen as the optimizer and Mean Squared Error was the loss function. For the SNN-Decoder, arctangent was applied as a surrogate function. A learning scheduler (cosine annealing schedule) was used after every epoch. For NN-Decoder, and ST-NN-Decoder, data was shuffled with batch size of 512 in training. Reset occurs at the beginning of each reach for membrane potential in SNN. For validation and testing, input is fed chronologically, and reset mechanisms only occur at the beginning of the evaluation.

\subsubsection{Decoder Performance Analysis}
The decoding performance is reported for the following configurations - The first EvFilter + Decoder and the second EvFilter-SPD + Decoder. The $\mathrm{R^2}$ scores obtained for the 'GT - Spike Times' were used as a baseline for comparison. From the results presented in Table \ref{tab:nndecoder_sota}, it is evident that the decoding performance for event-frame features using EvFilter/EvFilter-SPD is much better than what is obtained using the GT spike times provided in the dataset. 
Among the different variants of decoders studied in this work, SNN-Decoder ($\mathrm{R^2}:0.670-0.703 $) with inherent memory built into its model performs the best second only to an LSTM-Decoder ($\mathrm{R^2}:0.723-0.733 $) that takes the same Event-Frame inputs as NN-Decoder. Event-based features yielded superior performance for ST-NN-Decoder and SNN-Decoder, which may be attributed to the finer segmented timebins for ST-NN-Decoder, and streaming events used in the SNN-Decoder pipeline. The difference between the $\mathrm{R^2}$ scores of ST-NN-Decoder and SNN-Decoder is $\leq0.01$, indicating that a shallow ANN-based decoder can behave similar to SNN-Decoder if it can be provided with additional memory or temporal information. The better performance of the decoding pipeline with EvFilter-SPD compared to the one with GT spike times may be due to the presence of additional events that potentially correspond to spikes in the noise region, that may have not been captured by GT spike times.
Exploration of post-decoding processing that could potentially improve the $\mathrm{R^2}$ scores further is left as a future work.
While rEFH-dynamic \cite{NHP_HandReaching_dataset} shows slightly better results for GT spike times than the SNN/LSTM-Decoder (EvFilter-SPD input) as shown in Table \ref{tab:nndecoder_sota}, the former was trained on a much larger selection of datasets than the latter. The event-based input features with neural network decoders outperform other popular decoding algorithms including a variant of KF (SS-KF), linear regression, and a 2D-SNN decoder \cite{Neurobench}, operating on binned spike times, thereby validating the two main contributions of this work. 

\begin{table}[t]
\centering
\caption{Comparison of neural network based decoder studied in this work with event-based features using other popular decoders.}
\label{tab:nndecoder_sota}
\resizebox{0.4\textwidth}{!}{%
\begin{tabular}{|l|c|c|c|}
\hline
\multicolumn{1}{|c|}{\multirow{2}{*}{\textbf{Decoder}}} & \multirow{2}{*}{\boldmath{$\mathrm{T_{bin}}$}} & \multirow{2}{*}{\textbf{Input}} & \multirow{2}{*}{\boldmath{$\mathrm{R^2}$}} \\
\multicolumn{1}{|c|}{} &                         &               &        \\ \hline
\multirow{3}{*}{NN}    & \multirow{3}{*}{200}    & GT-Spike Time & 0.6560 \\ \cline{3-4} 
                       &                         & EvFilter     & 0.6685 \\ \cline{3-4} 
                       &                         & EvFilter-SPD & 0.6785 \\ \hline
\multirow{3}{*}{ST-NN} & \multirow{3}{*}{200}    & GT-Spike Time & 0.6823 \\ \cline{3-4} 
                       &                         & EvFilter     & 0.7070 \\ \cline{3-4} 
                       &                         & EvFilter-SPD & 0.7182 \\ \hline
\multirow{3}{*}{LSTM}  & \multirow{3}{*}{34}     & GT-Spike Time & 0.7233 \\ \cline{3-4} 
                       &                         & EvFilter     & 0.7272 \\ \cline{3-4} 
                       &                         & EvFilter-SPD & 0.7331 \\ \hline
\multirow{3}{*}{SNN}   & \multirow{3}{*}{Stream} & GT-Spike Time & 0.6703 \\ \cline{3-4} 
                       &                         & EvFilter     & 0.6980 \\ \cline{3-4} 
                       &                         & EvFilter-SPD & 0.7033 \\ \hline
\multirow{3}{*}{Linear regression}   & \multirow{3}{*}{300} & GT-Spike Time & 0.4789 \\ \cline{3-4} 
                       &                         & EvFilter     & 0.5083 \\ \cline{3-4} 
                       &                         & EvFilter-SPD & 0.5177 \\ \hline
\multirow{3}{*}{SS-KF}   & \multirow{3}{*}{300} & GT-Spike Time & 0.4246 \\ \cline{3-4} 
                       &                         & EvFilter     & 0.4833 \\ \cline{3-4} 
                       &                         & EvFilter-SPD & 0.4963 \\ \hline
rEFH dynamic \cite{NHP_HandReaching_dataset}           & 128                     & GT-Spike Time & 0.7301 \\ \hline
2D SNN \cite{Neurobench}                 & 128                     & GT-Spike Time & 0.5805 \\ \hline
UKF \cite{NHP_HandReaching_dataset}                    & 128                     & GT-Spike Time & 0.6135 \\ \hline
\end{tabular}%
}
\vspace{-0.5cm}
\end{table}

\subsection{Hardware Implications for Decoder}
\begin{figure}
    \centering
    \includegraphics[width=0.45\textwidth]{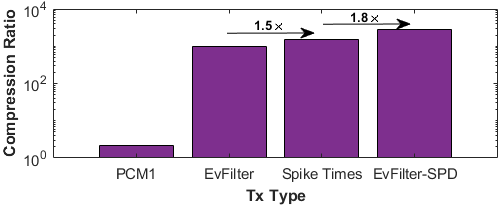}
    \caption{Compression ratio improvement with EvFilter/EvFilter-SPD.}
    \label{fig:EvFilter_CR}
    \vspace{-0.2cm}
\end{figure}

The Neurobench harness \cite{Neurobench} was used to obtain the metrics related to computes and memory to evaluate suitability for on-implant or on-wearable iBMI decoding. As shown in Table \ref{tab:nndecoder_computes_mem} unsurprisingly, the SNN-Decoder is the most resource-efficient since its event-based processing negates the need for multiplications, ensuring far fewer computations ($<1$K accumulations - ACs) compared to the other decoders ($3-16$K multiplications and accumulations - MACs). It was observed that  EvFilter and EvFilter-SPD reduce the number of events to be processed for decoding by about $192\times$ and $554\times$ respectively, compared to directly processing all the NCNS events for decoding as shown in Fig. \ref{fig:EvFilter_CR}.
\begin{table}[t]
\caption{Estimation of computes, memory access, and model size for the different neural network based decoders}
\label{tab:nndecoder_computes_mem}
\resizebox{0.49\textwidth}{!}{%
\begin{tabular}{|c|c|c|c|ccc|c|}
\hline
\multirow{2}{*}{\textbf{Decoder}} &
  \multirow{2}{*}{\textbf{\begin{tabular}[c]{@{}c@{}}$\mathrm{T_{bin}}$\\ (ms)\end{tabular}}} &
  \multirow{2}{*}{\textbf{Input}} &
  \multirow{2}{*}{\textbf{\begin{tabular}[c]{@{}c@{}}Activation\\ Sparsity\end{tabular}}} &
  \multicolumn{3}{c|}{\textbf{Computes}} &
  \multirow{2}{*}{\textbf{\begin{tabular}[c]{@{}c@{}}Model Size\\ (KB)\end{tabular}}} \\ \cline{5-7}
 &
   &
   &
   &
  \multicolumn{1}{c|}{\textbf{MACs}} &
  \multicolumn{1}{c|}{\textbf{ACs}} &
  \textbf{\begin{tabular}[c]{@{}c@{}}Memory\\ (Kb)\end{tabular}} &
   \\ \hline
\multirow{3}{*}{NN} &
  \multirow{3}{*}{200} &
  GT-Spike Time &
  0.6571 &
  \multicolumn{1}{c|}{3643.952} &
  \multicolumn{1}{c|}{0} &
  4.866 &
  \multirow{3}{*}{20.856} \\ \cline{3-7}
 &
   &
  EvFilter &
  0.6198 &
  \multicolumn{1}{c|}{3561.457} &
  \multicolumn{1}{c|}{0} &
  4.739 &
   \\ \cline{3-7}
 &
   &
  EvFilter-SPD &
  0.6225 &
  \multicolumn{1}{c|}{3561.084} &
  \multicolumn{1}{c|}{0} &
  4.739 &
   \\ \hline
\multirow{3}{*}{ST-NN} &
  \multirow{3}{*}{200} &
  GT-Spike Time &
  0.6649 &
  \multicolumn{1}{c|}{7121.185} &
  \multicolumn{1}{c|}{0} &
  10.041 &
  \multirow{3}{*}{94.584} \\ \cline{3-7}
 &
   &
  EvFilter &
  0.6285 &
  \multicolumn{1}{c|}{6205.122} &
  \multicolumn{1}{c|}{0} &
  8.659 &
   \\ \cline{3-7}
 &
   &
  EvFilter-SPD &
  0.6622 &
  \multicolumn{1}{c|}{6190.484} &
  \multicolumn{1}{c|}{0} &
  8.636 &
   \\ \hline
\multirow{3}{*}{LSTM} &
  \multirow{3}{*}{34} &
  GT-Spike Time &
  0 &
  \multicolumn{1}{c|}{16543.83} &
  \multicolumn{1}{c|}{0} &
  23.008 &
  \multirow{3}{*}{67.976} \\ \cline{3-7}
 &
   &
  EvFilter &
  0 &
  \multicolumn{1}{c|}{16543.83} &
  \multicolumn{1}{c|}{0} &
  23.008 &
   \\ \cline{3-7}
 &
   &
  EvFilter-SPD &
  0 &
  \multicolumn{1}{c|}{16543.83} &
  \multicolumn{1}{c|}{0} &
  23.008 &
   \\ \hline
\multirow{3}{*}{SNN} &
  \multirow{3}{*}{Stream} &
  GT-Spike Time &
  0.807 &
  \multicolumn{1}{c|}{0} &
  \multicolumn{1}{c|}{640.7854} &
  0.877 &
  \multirow{3}{*}{19.628} \\ \cline{3-7}
 &
   &
  EvFilter &
  0.7725 &
  \multicolumn{1}{c|}{0} &
  \multicolumn{1}{c|}{786.6741} &
  1.013 &
   \\ \cline{3-7}
 &
   &
  EvFilter-SPD &
  0.789 &
  \multicolumn{1}{c|}{0} &
  \multicolumn{1}{c|}{672.6938} &
  0.897 &
   \\ \hline
\end{tabular}%
}
\vspace{-0.5cm}
\end{table}
Finer temporal information in ST-NN-Decoder seems to improve performance compared to NN-Decoder, but it comes at the cost of higher computes and memory requirement (still $2.5\times<$ LSTM-Decoders). Overall, SNN-Decoder requires $5\times$, $9\times$, and $23\times$ fewer compute and memory than NN-, NN-ST-, and LSTM-Decoders, respectively. The proposed decoding pipeline introduced in this work can be realized efficiently on edge-based computing devices such as wearables, off-the-shelf COTS microprocessors, or even on ultra-low power neuromorphic processors such as Intel's Loihi \cite{Loihi}, DYNAP-SE \cite{DYNAP}, etc., for SNN-Decoder. While the estimates were obtained assuming 32-bit operations, the memory footprint of the models can be reduced significantly by decreasing the number of layers or hidden neurons, and by adopting model compression methods such as weight quantization. Further optimization of the decoder models for ultra-low power hardware realization is left for future work.

\section{Conclusion}
The proposed decoding pipeline for Neu-iBMI leverages sparse neural event data generated by NCNS, yielding high decoding performance with $\mathrm{R^2}=0.73$ using an LSTM-decoder after EvFilter-SPD. 
The SNN-Decoder achieves $\mathrm{R^2}=0.7$ with $23\times$ less memory and computations than the LSTM-Decoder, while the shallow NN-Decoder and its temporally segmented variant (ST-NN-Decoder) achieve $\mathrm{R^2}=0.67$ and $0.72$, respectively, using $2.5$-$4.8\times$ fewer resources.
The proposed pipeline eliminates additional steps typical of conventional iBMI systems, including signal recovery, spike sorting, and detection. This highlights the potential of combining Neu-iBMI with a hybrid event-frame iBMI decoder, supporting the realization of low-power, wireless Nx-iBMI systems with high channel counts to enhance capabilities and broaden applications.

\bibliographystyle{IEEEtran}
\bibliography{references}
\end{document}